\documentclass[runningheads]{llncs}
\usepackage{graphicx}

\usepackage{tikz}
\usepackage{comment}
\usepackage{amsmath,amssymb} 
\usepackage{color}
\usepackage{epsfig}
\usepackage{graphicx}
\usepackage{amsmath}
\usepackage{amssymb}
\usepackage{algorithm}
\usepackage{algpseudocode}
\usepackage{multirow}
\usepackage{caption} 
\usepackage{makecell}
\usepackage{subfigure}
\usepackage{blindtext}
\usepackage[pagebackref=true,breaklinks=true,letterpaper=true,colorlinks,bookmarks=false]{hyperref}
\usepackage{orcidlink}

\usepackage[accsupp]{axessibility}  

\usepackage{tikz}
\newcommand*\circled[1]{\tikz[baseline=(char.base)]{
            \node[shape=circle,draw,inner sep=0.5pt] (char) {#1};}}

\begin{document}
\pagestyle{headings}
\mainmatter
\def\ECCVSubNumber{2171}  

\newcommand{\systemName}{CPrune}

\title{\Large \bf \systemName: Compiler-Informed Model Pruning for Efficient Target-Aware DNN Execution} 


\titlerunning{CPrune: Compiler-Informed Model Pruning}
%
\author{Taeho Kim\inst{1}\orcidlink{0000-0001-7787-3035} \and
Yongin Kwon\inst{2}\orcidlink{0000-0003-2973-246X} \and
Jemin Lee\inst{2}\orcidlink{0000-0002-9332-3508} \and
Taeho Kim\inst{2}\orcidlink{0000-0002-5061-206X} \and
Sangtae Ha\inst{1}\orcidlink{0000-0001-5983-5430}}
\authorrunning{T. Kim et al.}
%
\institute{University of Colorado Boulder \\
\email{\{taeho.kim,sangtae.ha\}@colorado.edu} \and
Electronics and Telecommunications Research Institute \\
\email{\{yongin.kwon,leejaymin,taehokim\}@etri.re.kr}}
\maketitle

\begin{abstract}
Mobile devices run deep learning models for various purposes, such as image classification and speech recognition. Due to the resource constraints of mobile devices, researchers have focused on either making a lightweight deep neural network (DNN) model using model pruning or generating an efficient code using compiler optimization. Surprisingly, we found that the straightforward integration between model compression and compiler auto-tuning often does not produce the most efficient model for a target device. We propose \systemName, a compiler-informed model pruning for efficient target-aware DNN execution to support an application with a required target accuracy. \systemName\ makes a lightweight DNN model through informed pruning based on the structural information of subgraphs built during the compiler tuning process. Our experimental results show that \systemName\ increases the DNN execution speed up to 2.73$\times$ compared to the state-of-the-art TVM auto-tune while satisfying the accuracy requirement.
\end{abstract}
\footnotetext{Yongin Kwon is the corresponding author.}
\section{Introduction}
Deep neural networks (DNNs) have become increasingly popular for various applications like image processing and speech recognition. 
DNN requires heavy computation due to large and complex neural networks, so recent approaches proposed intelligent offloading using the cloud for resource-constrained mobile devices. Unfortunately, they often suffer from network overhead and disruptions, necessitating running DNNs directly on devices. Furthermore, making DNN inferences on devices is often desirable for user experience and privacy.

Several efforts have focused on improving DNN inference on mobile devices by considering energy efficiency, accuracy, and execution time. These efforts are primarily divided into two approaches: (1) making a lightweight DNN model using model compression and (2) generating optimized code for the target device using a DNN compiler. 
The model compression, for example, optimizes the neural networks by pruning, quantization, or neural architecture search (NAS). In particular, model pruning reduces the size of a model by removing non-critical or redundant neurons from a DNN model~\cite{liu2020pruning}, while quantization does it by reducing the precision of the datatype~\cite{liang2021pruning}. In addition, NAS optimizes the structure of a DNN model for maximizing performance~\cite{kyriakides2020introduction}. The DNN compiler, on the other hand, takes a DNN model and generates an optimized execution code for each target device.

\begin{figure}[t]
\begin{center}
 \includegraphics[width=0.8\linewidth]{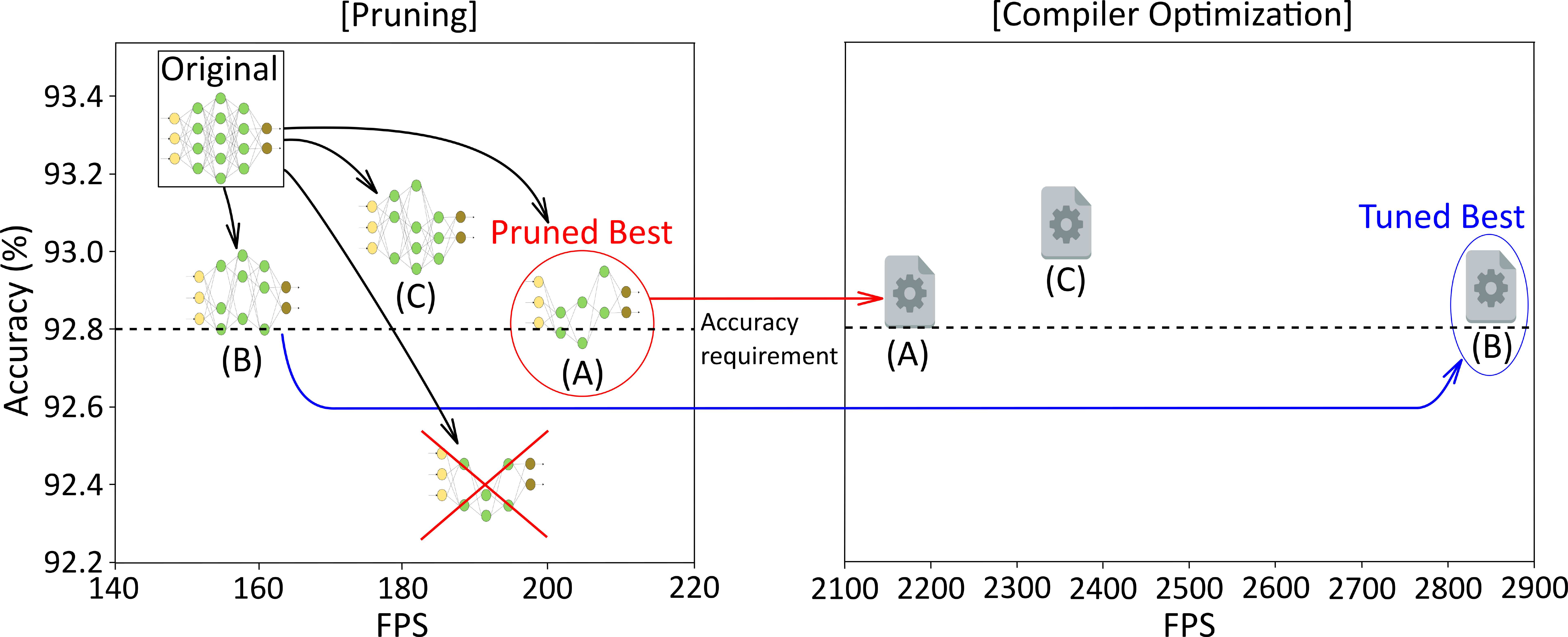}
\end{center}
  \caption{An experiment to find the fastest model whose accuracy is higher than 92.80\% accuracy for the target device among various pruned models of VGG-16 for CIFAR-10. We use the number of processed figures per second (FPS), given that indirect metrics like FLOPs cannot replace actual execution speed. The best-pruned model does not guarantee the best after compiler optimization. For example, the best model with pruning achieves only 2174 FPS while the suboptimal one obtains a higher 2857 FPS after compiler optimization. }
\label{fig:motivation}
\end{figure}

Efficient execution through a lightweight model is critical for many applications, such as AR/VR applications, autonomous driving~\cite{liu2021flexi}, drone or robot control~\cite{fang2018nestdnn}, medical health monitoring~\cite{kim2020epileptic}, malware detection~\cite{wei2017machine}, and user authentication such as Face ID~\cite{zhou2018echoprint,lu2019lip}. While one might think that the straightforward integration between model compression and compiler optimization could generate the most efficient code for a specific target device, this is not the case. 
Figure~\ref{fig:motivation} shows experimental results of direct integration between model compression and compiler optimization. As shown, the fastest model meeting the required accuracy found by pruning is often not the best model after compiler optimization, demanding joint optimization between them.  

Based on this observation, we propose \systemName, a new pruning technique for an efficient DNN execution for target devices that jointly considers model compression and compiler optimization. Instead of optimizing DNN compression and compiler optimization independently, leading to suboptimal optimization, \systemName\ exploits the information extracted during the compiler optimization process. In particular, \systemName\ uses the subgraph structures of a neural network and their execution times on the target device during compiler optimization to create an efficient target-aware pruned model fulfilling accuracy requirements. 
To the best of our knowledge, this is the first work using the information collected during the compiler optimization to create the most efficient target-aware compressed model meeting accuracy requirements. 

Our contributions are:
\begin{itemize}
    \item We report from our experiments that the fastest model that meets the required accuracy found by pruning is often not the best model after compiler optimization, necessitating joint optimization.
    \item We propose \systemName, which incorporates the information extracted during the compiler optimization process into creating a target-oriented compressed model fulfilling accuracy requirements. This information also significantly reduces the search space for parameter tuning such that  \systemName\ can make a compressed model substantially faster than NetAdapt~\cite{yang2018netadapt}, the state-of-the-art hardware-aware model compression framework. 
    \item Our experimental results show that \systemName\ achieves target-oriented performance improvement and increases the speed up to 2.73$\times$ compared to the de facto DNN compiler framework TVM auto-tune while satisfying the accuracy requirement.
\end{itemize}

While CPrune's approach is generally applicable, we can utilize CPrune for computer vision tasks requiring fast processing but a slight reduction in accuracy is acceptable, such as object detection and image classification for autonomous driving and biosignal image processing for seizure detection on a mobile device. 
We implement \systemName\ on top of an open deep learning compiler stack Apache TVM~\cite{tvmGithub} 
and Microsoft NNI~\cite{nniGithub}. 
Our source code can be found at~\href{https://github.com/taehokim20/CPrune}{https://github.c \\ om/taehokim20/CPrune}.
\section{Related Works}
\subsection{Model Compression}\label{related_1}

Model compression reduces the model size and improves speed while meeting the performance requirements: pruning compresses a model while NAS creates a much lighter model. There are also approaches using the performance measurements on real hardware as feedback for model compression. 

\begin{figure}[t]
\begin{center}
  \includegraphics[width=0.75\linewidth]{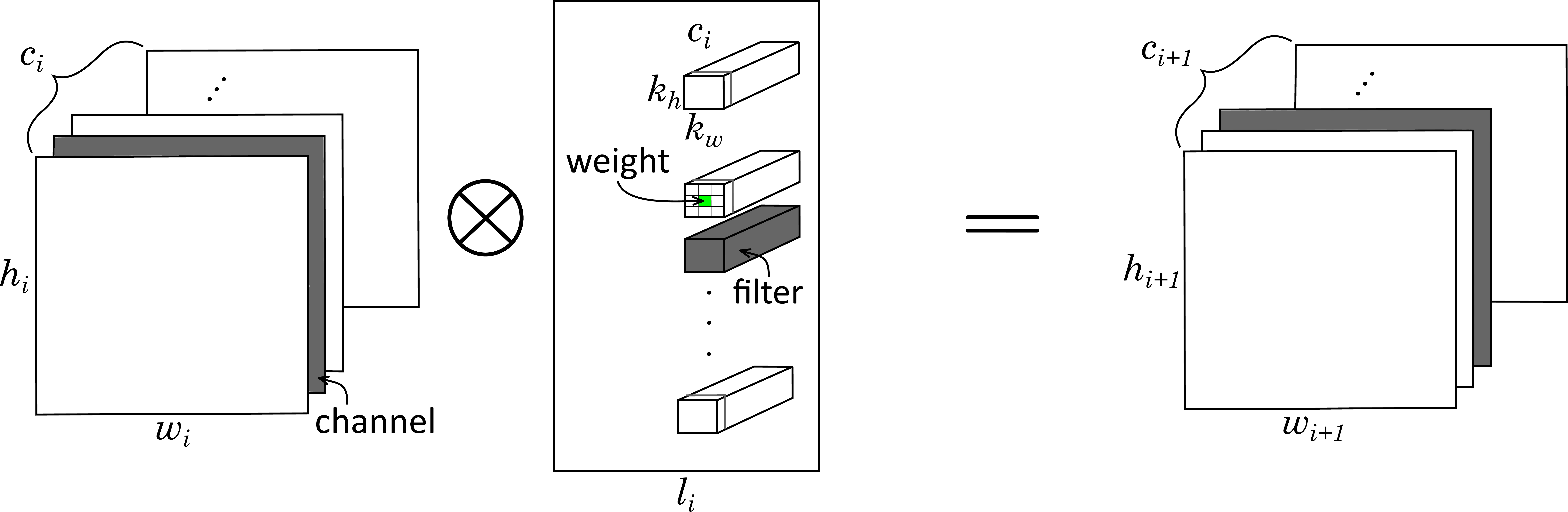}
\end{center}
  \caption{Structured pruning removes channels or filters to reduce the model's overall size. In contrast, non-structured pruning selectively prunes less important weights to minimize the sizes without impacting accuracy loss.}
\label{fig:prunings}
\end{figure}

\textbf{Pruning.} Pruning removes non-critical, redundant neurons from a DNN model~\cite{liu2020pruning}. Figure~\ref{fig:prunings} shows an example of a convolution operation in DNN. $c_i$ is the number of input channels, each of which comprises height ($h_i$) and width ($w_i$) of the input. After convolving $c_i$ input channels 
with filters of layer $l_i$, the input becomes abstracted to a feature map with output channels $c_{i+1}$, where $c_{i+1}$ is determined by the number of filters in layer $l_i$. 
Structured pruning~\cite{he2018amc,he2017channel,li2016pruning,wang2021convolutional,zhao2019variational} removes filters or channels according to various algorithms to judge the redundancy of weights. For example, removing a few input channels $c_i$ reduces the overall sizes of the input and filter shape, and pruning a filter reduces the sizes because it eliminates a channel in the next layer (both cases are highlighted as shaded regions in Figure~\ref{fig:prunings}). On the other hand, non-structured pruning~\cite{guo2016dynamic,zhang2018systematic} selects individual weights to prune. As a result, they achieve very high compression without much accuracy loss even though it is hard to gain as much as speedups of structured ones~\cite{trevor2020sparse,zhang2020sparse,ma2020pconv}.

\textbf{NAS.} NAS optimizes the structure of a DNN model for maximizing performance~\cite{elsken2019neural}. It uses various search strategies to explore the search space of neural architectures, such as random search, Bayesian optimization~\cite{shahriari2015taking,white2019bananas}, evolutionary method~\cite{periaux2015evolutionary,chen2019renas}, reinforcement learning (RL)~\cite{zoph2016neural}, and gradient-based methods~\cite{elsken2019neural,liu2018darts}.

\textbf{Hardware-aware model compression.} 
Recently, several approaches have utilized performance measurements on real hardware to make a compressed model~\cite{li2021npas,tan2019mnasnet,wu2019fbnet,yang2018netadapt,yang2021netadaptv2}. Especially, NetAdapt~\cite{yang2018netadapt} points out that indirect metrics (e.g., the number of MACs or weights) may not necessarily reduce the direct metrics (e.g., execution time and energy consumption)~\cite{yang2017designing,yu2017scalpel}, so they incorporate performance metrics of target hardware into its pruning adaptation algorithm. 

Note that our approach \systemName\ is different from NetAdapt in that it jointly considers model compression and compiler optimization.

\subsection{DNN Compilers with Auto-Tuning}

Various DNN compilers generate code for DNN models on different hardware architectures~\cite{chen2018tvm,lattner2021mlir,rotem2018glow}. For example, TVM~\cite{chen2018tvm} translates the input DNN models into the intermediate representation (IR) during the compilation process~\cite{roesch2018relay}. After performing hardware-independent optimizations with the IR, it translates high-level IR into low-level IR, which provides interfaces to tune the computation and memory access. 
In addition, auto-tuning approaches such as AutoTVM~\cite{chen2018learning} and AutoScheduler~\cite{zheng2020ansor} optimize various low-level hardware-dependent parameters since the optimal parameters vary depending on the DNN operations and the target hardware. 
\section{\systemName: Compiler-Informed Model Pruning}
Our motivation starts from a question: \textit{Can an optimally pruned model guarantee the best performance after compiler optimization for a specific device?} To check this question, we prune VGG-16~\cite{simonyan2014very} randomly and create 20 different models. We use the CIFAR-10~\cite{krizhevsky2009learning} dataset, and the original model's accuracy is 93.29\%. We check if the model showing the highest number of processed figures per second (FPS) with an accuracy higher than 92.80\% would show the highest FPS after compiler optimization. We use FPS, given that indirect metrics like FLOPs cannot replace actual execution speed. For this experiment, we run and measure the FPS on a PC with an NVIDIA RTX-3080 GPU.

As shown in Figure~\ref{fig:motivation}, when we measure the FPS of each pruned model, the fastest model that meets the accuracy requirement is model A (92.85\%, 205 FPS). However, after compiler optimization on the pruned models, the fastest model is model B (92.86\%, 2857 FPS). It is different from what we expect: the most efficient model after pruning should be the most efficient one after compiler optimization. We also find no strong correlation between the performance of pruned models before and after compiler optimization, emphasizing the need for joint optimization between model compression and compiler optimization.  

Inspired by this finding, we design a new pruning scheme, \systemName, that generates an efficient DNN execution model by pruning a model based on the extracted information during compiler optimization for the target device. Below we present the details of our approach. 

\begin{figure*}[t]
\begin{center}
  \includegraphics[width=0.99\linewidth]{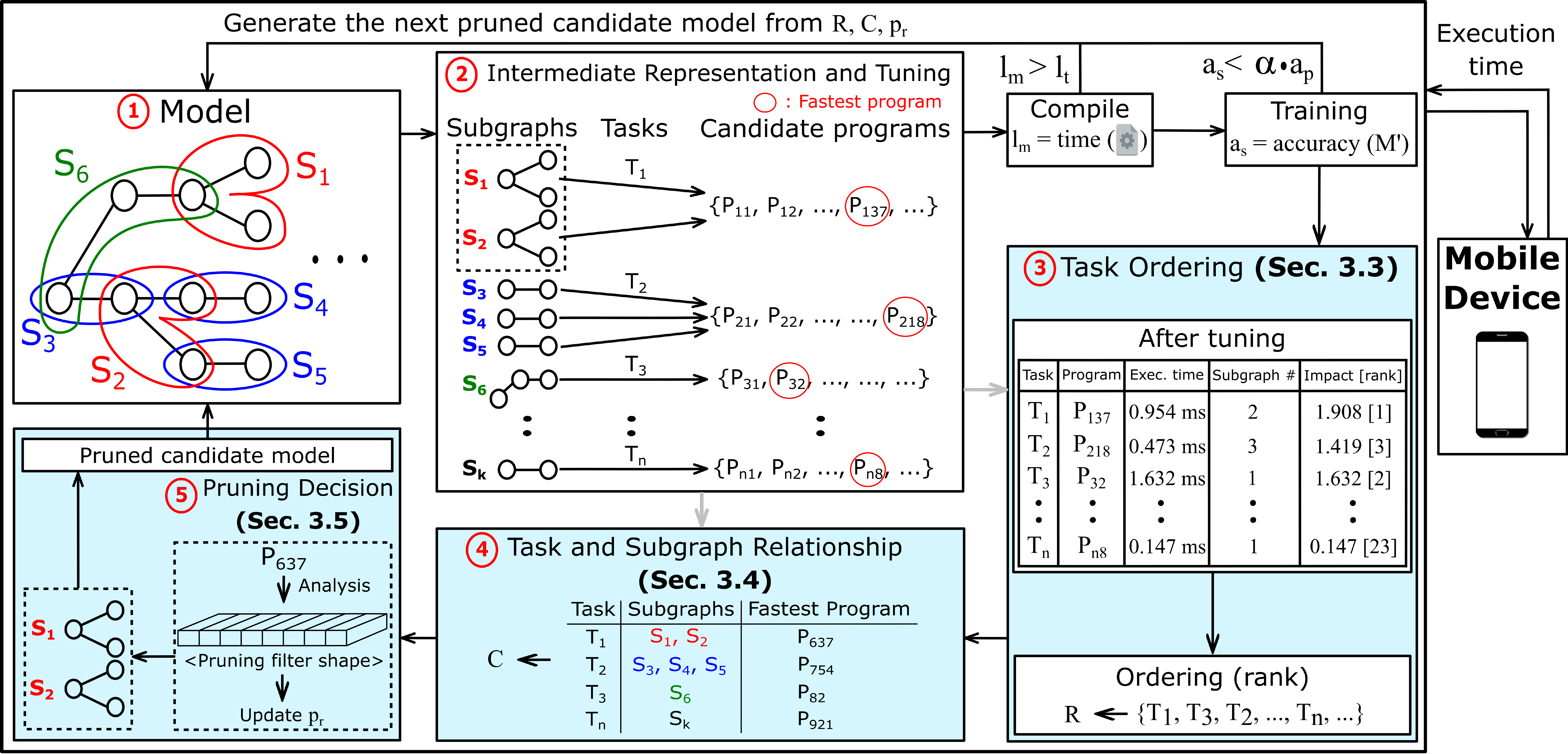}
\end{center}
  \caption{\systemName\ overview. \systemName\ generates the fastest DNN execution model on the target device by using the structural information extracted during compiler optimization and the on-device performance of a DNN model.}
\label{fig:overview}
\end{figure*}

\subsection{\systemName\ Overview}

The goal of \systemName\ is to find and prune neurons that primarily impact the execution of the DNN model on the target device while meeting the accuracy requirements. This informed pruning effectively prevents the issue of the best pruning model not being the best on the target device after compilation. 

Figure~\ref{fig:overview} depicts how \systemName\ leverages the information collected during compiler optimization to prune neurons from the DNN model effectively. The shaded boxes indicate what \systemName\ adds to the existing compiler optimization framework. A DNN model consists of many convolutional layers, each representing a subgraph (\circled{1}). A subgraph is assigned to a task, and two different subgraphs can point to the same task. Then, a DNN compiler creates numerous intermediate representations for each task and selects the fastest program on the target device (\circled{2}). After compiling the aggregate of the fastest programs comprising a DNN model, \systemName\ checks if the model meets the minimum execution and accuracy requirements. 
After that, \systemName\ sorts tasks based on their execution times in descending order to find the most efficient model with further pruning. Since pruning a task with a longer execution time could significantly reduce a model's execution time, \systemName\ selects the most time-consuming task as a candidate for further pruning (\circled{3}).
\systemName\ now needs to know which subgraph(s) is associated with this task as pruning candidates. CPrune also store the fastest program for each task. 
For this purpose, \systemName\ builds a table keeping the relationship among tasks, subgraphs, and programs (\circled{4}). Finally, \systemName\ prunes subgraphs of the selected task while ensuring their code structures follow the structure of the fastest program of that task (\circled{5}). Since the computation structure impacts the execution time, preserving the same computation structure after pruning is critical for efficient pruning. This process continues to make the most efficient pruned DNN model satisfying the accuracy requirement.

In the following subsections, we describe the the details of an algorithm (Section~\ref{algo}), task ordering (Section~\ref{task_ordering}), task and subgraph relationship table (Section~\ref{subgraph_task_connection}) and pruning decision (Section~\ref{pruning_filters}).   

\subsection{\systemName \ Algorithm}\label{algo}

\systemName\ prunes a DNN model gradually and iteratively. In each iteration, \systemName\ analyzes subgraphs of the model and creates relationships between subgraphs and each task. Then, it takes a pre-trained model $M$ and the minimum accuracy requirement $a_g$ and returns the efficient target-aware DNN executable program, as shown in Algorithm~\ref{alg:cap}. The computational complexity of Algorithm 1 and a table that summarizes variables used are in the Supplementary Materials.

\textbf{Initialization step (Line 1).} It initializes tuning-related parameters, including the pruning rate of each subgraph $p_r$, the target execution time of the following iteration $l_t$, and the short-term accuracy of the previous best model $a_p$ in Algorithm~\ref{alg:cap}. It also initializes a task/subgraph table $C$, storing the relationship among tasks, subgraphs, and fastest programs and the list of tasks prioritized by tuning $R$.

\textbf{Main step: pruning based on the compiler optimization (Line 2-16).} 
In each pruning iteration (the while loop in Algorithm~\ref{alg:cap}) of this step, \systemName\ generates a pruned candidate model (\circled{1} in Figure~\ref{fig:overview}) and moves to the intermediate representation (IR) and tuning stage (\circled{2}).
When a task in order of $R$ is picked up (\circled{3}), \systemName\ extracts associated subgraphs and the fastest program of the task from $C$ (\circled{4}). It decides the number of filters to prune by analyzing the arrangement of filters of the program and prunes filters of the subgraphs to create a pruned candidate model $M'$ (\circled{5}, details in Section~\ref{pruning_filters}). 
\systemName\ uses the relationship between subgraphs of the model $M'$ and each task and creates a task/subgraph table $C'$. It also puts the fastest program of each task in $C'$ (details in Section~\ref{subgraph_task_connection}).
Furthermore, the IR and tuning process maintains a candidate list of prioritized tasks $R'$ for the next iteration (details in Section~\ref{task_ordering}).
If the measured execution time of $M'$ ($l_m$) is less than $l_t$, it trains $M'$ shortly and measures the short-term accuracy $a_s$.
When $a_s$ meets the requirement of the current iteration, it updates parameters and goes to the next iteration. $\alpha$ is the ratio to represent the minimum allowable accuracy after pruning, and $\beta$ is the ratio to define the target execution time of the next pruning iteration. 

\begin{algorithm}[t]
\caption{\systemName\ Algorithm}\label{alg:cap}
\textbf{Input}: Pre-trained model $M$ and $a_g$ \\
\textbf{Output}: An efficient DNN executable file of $M$
\begin{algorithmic}[1]
\State Tune $M$ and initialize $p_r$, $l_t$, $a_p$, $C$ and $R$
\While{$a_p > a_g$ \textbf{and} $R \neq \{ \}$}
    \For{$r$ in $R$}
        \State For $r$, obtain associated subgraphs ($S$) and the fastest program ($P$) from $C$
        \State Update pruning rate ($p_r$) by analyzing the structure of $P$
        \State Create a pruned candidate model ($M'$) by pruning $S$ with pruning rate $p_r$
        \State Extract tasks from $M'$ and create a candidate task/subgraph table ($C'$)
        \State Tune these tasks and create a candidate list of prioritized tasks ($R'$)
        \State Compile the aggregate of tuned tasks and measure $l_m$
        \State Go line 3 if $l_m \geq l_t$
        \State Short-term train $M'$ and measure $a_s$
        \State Remove $r$ from $R$ and go line 3 if $a_s < \alpha \cdot a_p$
        \State $M \gets M'$, $R \gets R'$, $C \gets C'$, update $p_r$, $l_t = \beta \cdot l_m$, $a_p = a_s$
        \State \textbf{break}
    \EndFor
\EndWhile
\State Final long-term training, tuning and compilation for $M$
\end{algorithmic}
\end{algorithm}

If $l_m$ is greater than or equal to $l_t$, the system selects the next task from $R$ to create the next pruned candidate model. 
Once $a_s$ is less than the target accuracy of the current iteration $(\alpha \cdot a_p$), \systemName\ removes the current task from $R$ and does not consider this task as a candidate for pruning in the rest iterations.

\textbf{Final step: final training and tuning (Line 17).}
If there are no more tasks in $R$ that can be pruned while meeting the accuracy requirement, \systemName\ progresses to the final step. In this step, \systemName\ trains and tunes the final model to achieve optimal accuracy and execution time.

\subsection{Task Ordering} \label{task_ordering} 

\systemName\ needs to select the most promising task of the current model $M$ for pruning. Therefore, \systemName\ sorts tasks according to the pruning impact (task's execution time $\times$ number of subgraphs associated with the task). 
The higher the pruning impact, the more likely the task satisfies the current iteration's target execution time $l_t$. 

The rightmost shaded box in Figure~\ref{fig:overview} shows the result of task ordering as an example. The pruning impact of $T_1$ is $0.954 \times 2 = 1.908$, $T_2$ is $0.473 \times 3 = 1.419$, $T_3$ is $1.632 \times 1 = 1.632$, and so on. After \systemName\ performs task ordering including the results of other tasks, the ordered list $R'$ becomes \{$T_1$, $T_3$, $T_2$, ..., $T_n$, ...\}. In this example, the subgraphs associated to $T_{1}$ will be pruned first. If this choice does not satisfy the execution time and accuracy requirements, $T_{3}$ and then $T_{2}$ will be selected for the next pruning candidate.

\subsection{Task and Subgraph Relationship} \label{subgraph_task_connection}
\begin{figure}[t]
\begin{center}
    \includegraphics[width=0.95\linewidth]{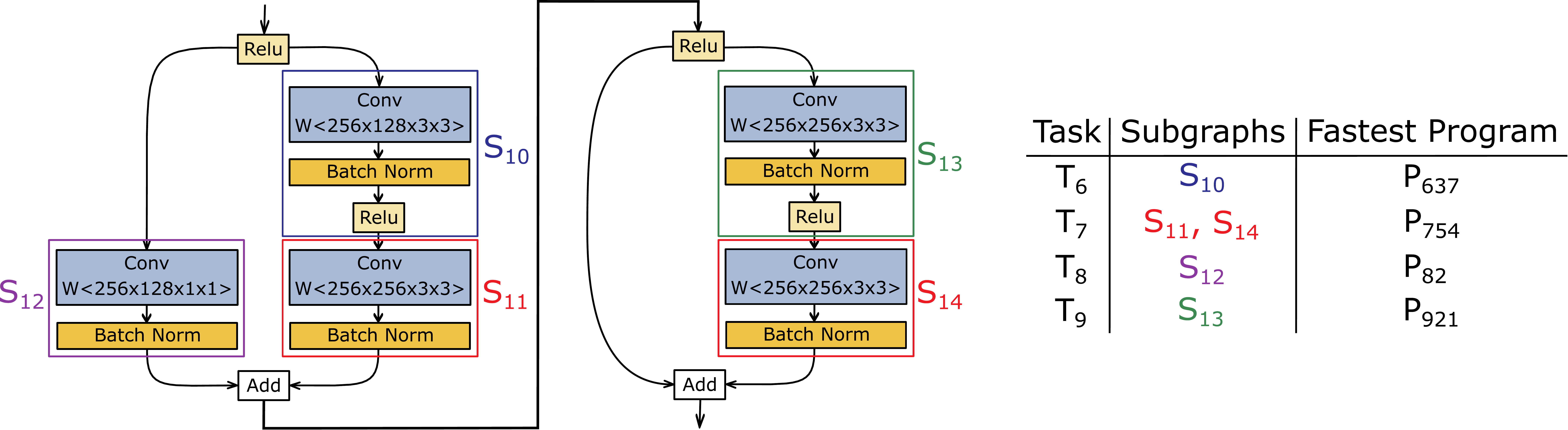}
\end{center}
  \caption{The graph analysis of \systemName \ for the ResNet-18 model}
\label{fig:table2_graph}
\end{figure}

\systemName\ creates a table storing the relationship among tasks, subgraphs, and fast programs to determine which subgraphs are pruned and which program is analyzed when selecting a task in the next pruning iteration. \systemName\ analyzes a pruned candidate model by checking the layer connectivity, weight shape, Rectified Linear Unit (ReLU), and Batch Normalization (BN). Figure~\ref{fig:table2_graph} shows \systemName's graph analysis for part of the ResNet-18 model~\cite{torchvisionModels}. As ResNet-18 consists of multiple convolutions of the same shape, the compiler partitions the large computational graph of a DNN into multiple subgraphs, each of which can be associated with other subgraphs~\cite{roesch2018relay}. For example, subgraphs $S_{11}$ and $S_{14}$ are connected to the same task $T_7$ due to the same properties of BN and input shapes. On the other hand, subgraphs $S_{10}$, $S_{12}$, and $S_{13}$ are connected to different tasks, respectively. In addition, each task is connected to its fastest program during the tuning.

\begin{figure}[t]
\begin{center}
    \includegraphics[width=1\linewidth]{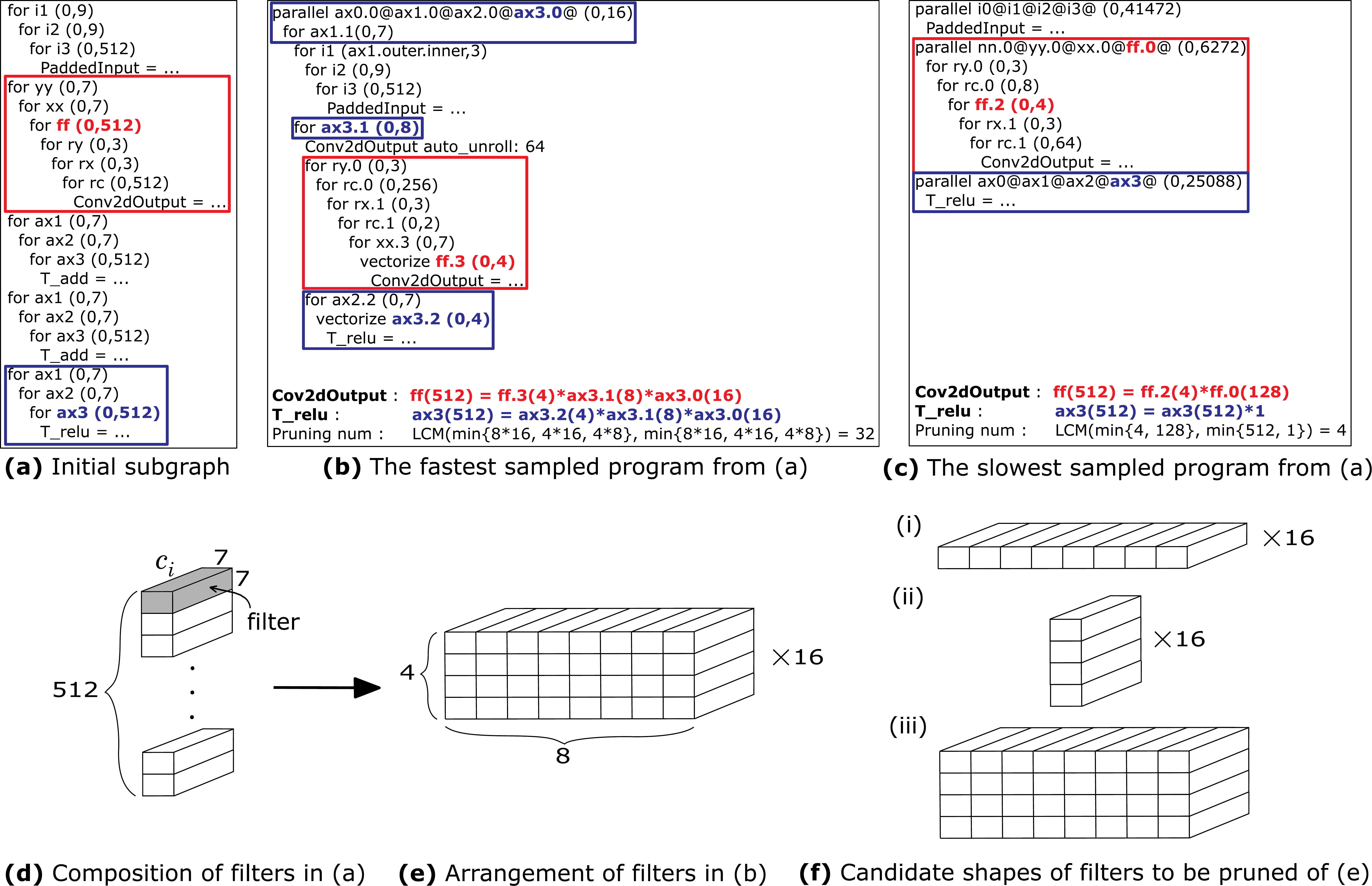}
\end{center}
  \caption{Sampled program comparison extracted from one subgraph among several subgraphs of ResNet-18. $c_i$ is the number of input channels.}
\label{fig:sampled_program}
\end{figure}

\subsection{Pruning Decision} \label{pruning_filters}

\systemName\ prunes filters while maintaining the structure of the fastest program created during the IR tuning process of compiler optimization. Figures~\ref{fig:sampled_program} (b) and (c) compare the structures of the fastest and slowest programs generated from the same subgraph in Figure~\ref{fig:sampled_program} (a). As we can see, the fastest program Figure~\ref{fig:sampled_program} (b) allows a subgraph to perform parallel processing of computations most efficiently. In contrast, the slowest program Figure~\ref{fig:sampled_program} (c) shows the most inefficient parallel processing of computations. If \systemName\ prunes four filters suitable for maintaining the structure of the program Figure~\ref{fig:sampled_program} (c), a DNN compiler may generate an inefficient program for the corresponding subgraph of a DNN model. Instead, pruning the subgraph to follow the program structure of the fastest program would lead a DNN compiler to generate an efficient program.
In addition, since the execution time of the convolution layer increases in a step pattern rather than linear with the number of filters~\cite{tang2021bridge}, pruning an insufficient number of filters would only increase the tuning time without improving FPS.

In particular, \systemName\ splits a given model into multiple subgraphs and determines the number of filters to prune for each subgraph. We elaborate on calculating the number of filters to be pruned by taking one subgraph as an example, as shown in Figure~\ref{fig:sampled_program} (a). We can see that the total number of filters is 512, and the kernel shape is 7 $\times$ 7 like Figure~\ref{fig:sampled_program} (d) by checking the 2D convolution layer (Conv2d) and the ReLU activation function layer (T\_relu) in Figure~\ref{fig:sampled_program} (a). Iterators related to the number of filters are \texttt{ff} and \texttt{ax3}. With this information, \systemName\ checks how these iterators are further split or merged. For example, Figure~\ref{fig:sampled_program} (b) shows that 512 of \texttt{ff} in Figure~\ref{fig:sampled_program} (a) is converted to 4$\times$8$\times$16 by the argument \texttt{ff.3}. 512 of \texttt{ax3} is also converted to 4$\times$8$\times$16. Therefore, both iterators arrange filters like Figure~\ref{fig:sampled_program} (e).

Finally, \systemName\ calculates the minimum number of filters to prune based on the arrangement shapes of filters obtained from the related iterators. If \systemName\ reduces the number in width, height, or depth in the arrangement of filters of Figure~\ref{fig:sampled_program} (e), the subgraph can maintain the structure of Figure~\ref{fig:sampled_program} (b) after pruning the filters. However, if \systemName\ prunes many filters at once, it cannot achieve optimal pruning due to the high difference from the performance goal in the current pruning iteration. Therefore, we prune filters by the step size to improve performance by pruning in the convolution layer~\cite{tang2021bridge}. Let $L_1=\{a_1, a_2, ..., a_m\}$ be the set of product combinations of the first iterator, and $L_2=\{b_1, b_2, ..., b_n\}$ be the set of product combinations of the second iterator. \systemName\ determines the least common multiple (LCM) for the two pruning numbers as the minimum number of pruning filters using $L_1$ and $L_2$.

\begin{align*}
    LCM(\min_{l \in L_1} \frac{\prod_{k=1}^m a_k}{l}, \min_{l \in L_2} \frac{\prod_{k=1}^n b_k}{l})
    = LCM(\frac{\prod_{k=1}^m a_k}{\max_{l \in L_1} l}, \frac{\prod_{k=1}^n b_k}{\max_{l \in L_2} l})
\end{align*}

\noindent
For example, we calculate the minimum number of filters that can be pruned while maintaining the program structure in Figure~\ref{fig:sampled_program} (b) and Figure~\ref{fig:sampled_program} (c), respectively. In Figure~\ref{fig:sampled_program} (b), the product combination of iterators \texttt{ff} and \texttt{ax3} is the same as 4$\times$8$\times$16. \systemName\ can prune 8$\times$16, 4$\times$16, or 4$\times$8 filters like Figure~\ref{fig:sampled_program} (f) while maintaining the product combination. Thus, the minimum number of pruning filters of Figure~\ref{fig:sampled_program} (b) is $LCM(\min \{8\times16, 4\times16, 4\times8\}, \min \{8\times16, 4\times16, 4\times8\}) = 
32$. On the other hand, in Figure~\ref{fig:sampled_program} (c), \texttt{ff} and \texttt{ax3} are converted to 4$\times$128 and 512$\times$1, respectively. Therefore, the minimum number of pruning filters of Figure~\ref{fig:sampled_program} (c) is $LCM(\min \{4, 128\}, \min \{512, 1\}) = 
4$.

After determining the number of filters to be pruned, \systemName\ decides which filters to prune. It calculates the sum of each filter's absolute weights (i.e., $l_1$-norm) and prunes filters starting with the smallest sum~\cite{nniGithub,li2016pruning}.\footnote{Using other metrics can improve the performance as well.} \systemName\ also prunes the input channels of the next layer by the determined number as described in Section~\ref{related_1} to maintain the connection consistency. In the case of the shortcut part in a ResNet model, we prune the same output channels using the dependency-aware mode supported by Microsoft NNI~\cite{nniGithub}.

\section{Experiments}

\subsection{Experimental Setup}
In this section, we carry out various experiments to evaluate the performance of \systemName. We conduct experiments over different pre-trained DNN models, including ResNet-18~\cite{he2016deep}, MobileNetV2~\cite{sandler2018mobilenetv2}, and MnasNet1.0~\cite{tan2019mnasnet}, with ImageNet~\cite{krizhevsky2012imagenet} or CIFAR-10~\cite{krizhevsky2009learning} datasets on various resource-constrained mobile devices (Samsung Galaxy S8 (Kryo 280 CPU), S9 (Kryo 385 CPU and Mali-G72 GPU), S20+ (Kryo 585 CPU), or Google Pixel 3 XL (Kryo 385 CPU)). We also use multiple host PCs with NVIDIA GeForce RTX 1080 Ti or 2080 Ti. All pruned models are optimized by stochastic gradient descent (SGD)~\cite{robbins1951stochastic}. The number of training epochs considered varies according to the choice of the dataset and the specific phase of training (i.e., either short-term or final training). For the CIFAR-10 dataset, the short-term and final training epochs are 5 and 100, respectively. For the ImageNet dataset, the training epochs are one-fifth of CIFAR-10 due to the enormous data size. Due to space limitations, the details of finding reasonable $\alpha$ and $\beta$ values and additional experiments about the impact of tuning and selective search are in the Supplementary Materials.

\begin{figure}[t]
    \begin{center}
            \subfigure[Short-term accuracy]{
                \includegraphics[width=0.36\columnwidth]{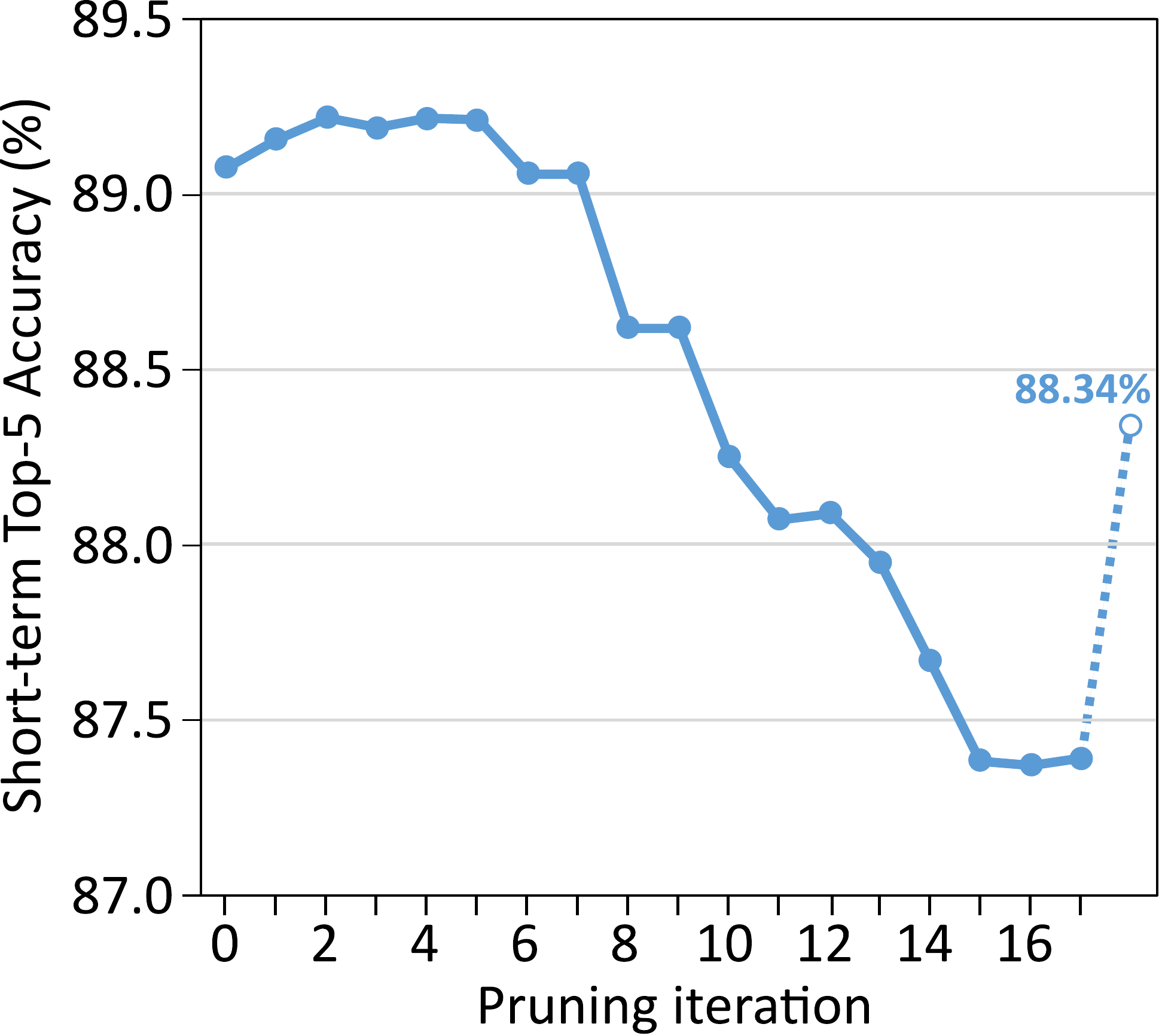}
            }
            \subfigure[FPS increase rate]{
                \includegraphics[width=0.35\columnwidth]{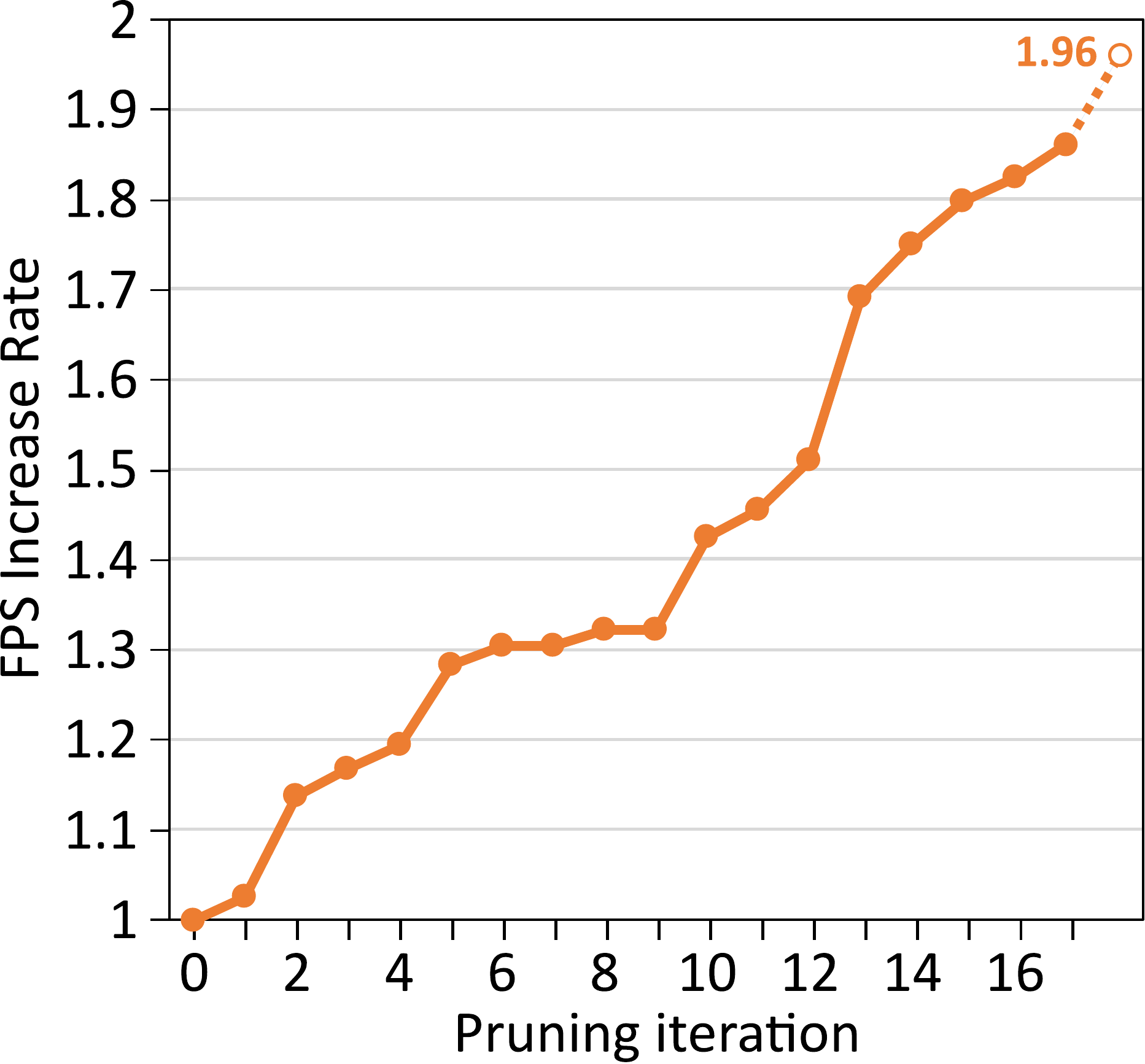}
            }
    \end{center}
    \caption{The FPS increase rate and short-term accuracy during the iterative \systemName\ process compared to using only TVM auto-tune. After completing the compiler-informed pruning process, \systemName\ performs final long-term training and tuning (Line 17 of Algorithm~\ref{alg:cap}). Therefore, we marked the final results with dotted lines to distinguish them from other short-term measurements.}
    \label{fig:process}
\end{figure}

\subsection{Overall Performance}
This experiment shows \systemName's FPS and short-term accuracy during the iterative pruning process compared to the case of using only a DNN compiler and its auto-tuning. For this experiment, we select TVM auto-tune~\cite{zheng2020ansor} for the DNN compiler and use the  ResNet-18 model on Kyro 385 CPU using the ImageNet dataset. 
Figure~\ref{fig:process} shows the results. In each pruning iteration, \systemName\ modifies the current model by pruning selected subgraphs. For any given pruning iteration, if a pruned candidate model satisfies the condition of the given iteration ($l_t$ and $\alpha \cdot a_p$), the candidate model is selected, and the iteration progresses. 
The final accuracy is 88.34\%, and the FPS increase rate is 1.96 times faster than using TVM auto-tune alone. We emphasize that our \systemName's FPS increase rate is around two times that obtained by only using the TVM auto-tune, while the final accuracy is within tolerable limits. Note that \systemName\ can stop its pruning around the $6^{th}$ pruning iteration if the accuracy requirement is more than 89\%. Then the FPS increase rate is 1.3 times faster than TVM auto-tune. 

In a practical scenario, user applications can provide the accuracy and execution time requirements to \systemName.

\subsection{Performance on Different Target Devices}

\begin{figure}[t]
\begin{center}
 \includegraphics[width=0.6\linewidth]{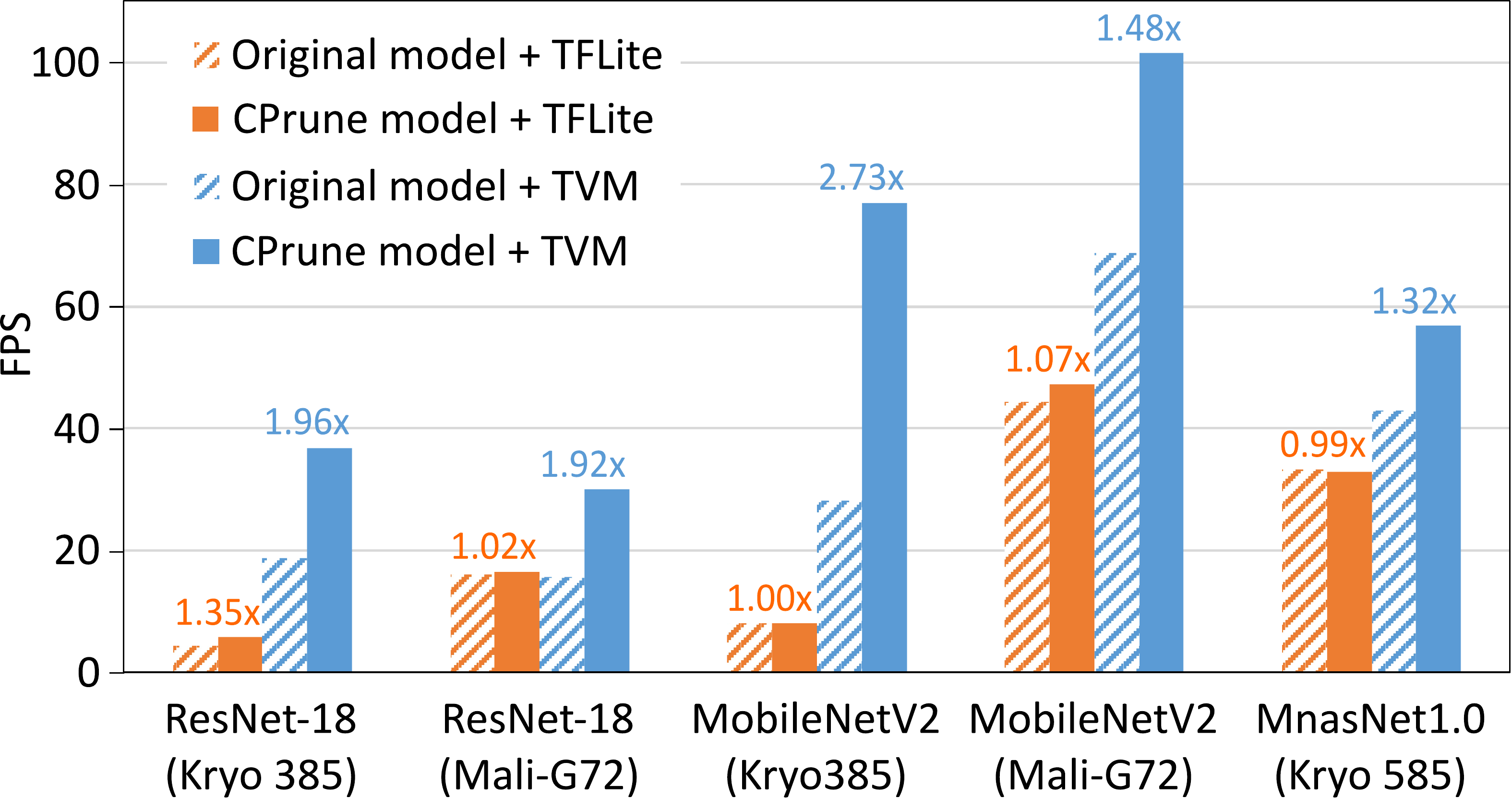}
\end{center}
  \caption{Comparison of FPS when we execute \systemName\ model with the target compiler framework TVM and a target-agnostic deep learning framework TFLite.}
\label{fig:target_compiler_comparison}
\end{figure}

\begin{figure}[t]
\begin{center}
 \includegraphics[width=0.7\linewidth]{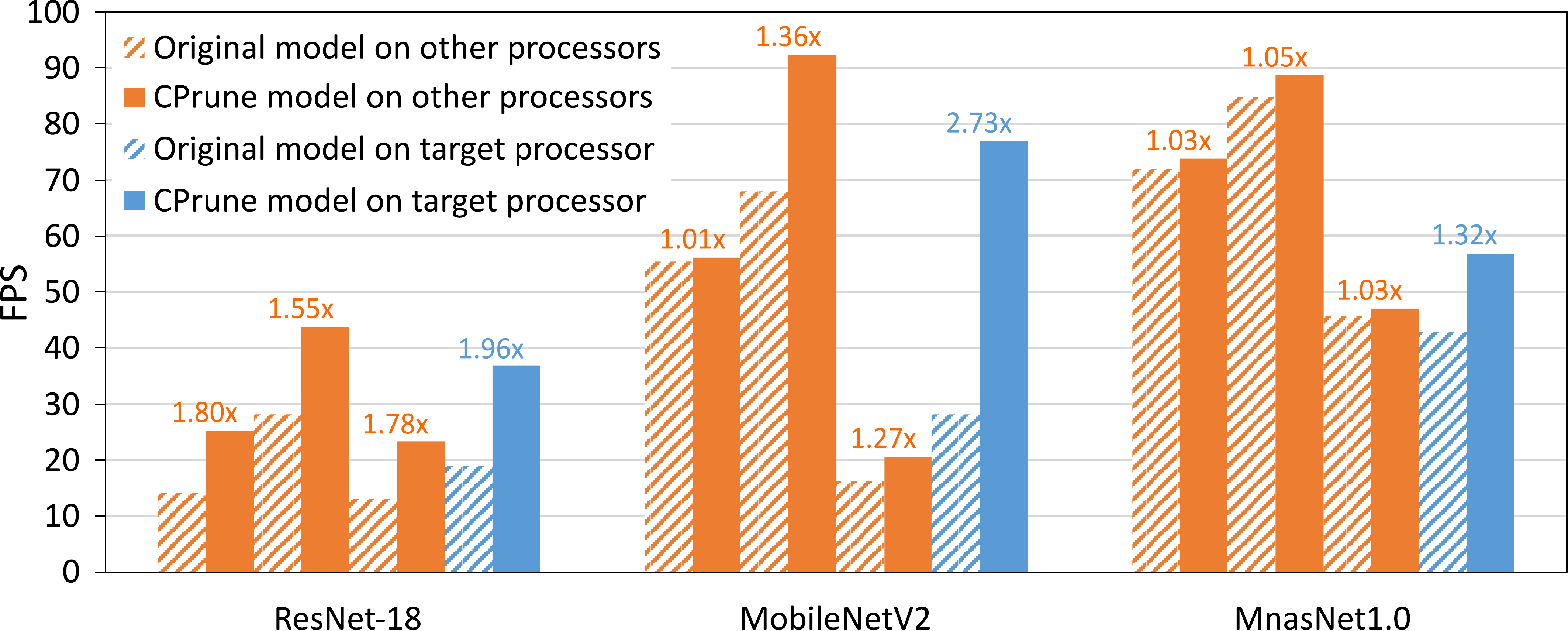}
\end{center}
  \caption{Comparison of FPS when we execute \systemName\ model on a target processor and other processors.}
\label{fig:target_device_comparison}
\end{figure}

This section evaluates \systemName's performance on different target devices. For this experiment, we integrate \systemName\ with the TVM compiler to compare its performance on different target devices. We also convert the final pruned model to a widely used TFLite executable for performance comparison. We compare the FPS of different DNN models using the ImageNet dataset on mobile CPU (Kryo 385 or Kryo 585) and GPU (Mali-G72), as shown in Figure~\ref{fig:target_compiler_comparison}. \systemName, along with the target compiler framework, shows a significantly higher FPS than the cases of running only a DNN compiler (e.g., TVM) and a target agnostic DNN library (e.g., TFLite).
Furthermore, regardless of the type of processor and model, the FPS increase rate when executing the \systemName\ model on a target processor is significantly higher than when we run it on other processors, as shown in Figure~\ref{fig:target_device_comparison}.

\subsection{Comparison with Other Pruning Schemes}

This section compares \systemName\ with other pruning schemes on different mobile platforms. We also integrate \systemName\ into TVM for a fair comparison. Table~\ref{tab:performance_test_imagenet} shows the results. \systemName\ shows a higher FPS than the model-based pruning models (e.g., PQF~\cite{martinez2021permute}, FPGM~\cite{he2019filter}, AMC~\cite{he2018amc}).
\systemName\ also shows similar or better performance than the hardware-aware pruning model (e.g., NetAdapt~\cite{yang2018netadapt}) with TVM. 
It also shows that indirect metrics such as FLOPS and parameters do not fully reflect actual performance. While FLOPS is a suitable indirect measure of the extent of compression obtained during the pruning process, FPS suitably reflects the pruning gains in terms of execution times or speeds.

\begin{table*}[t]
\caption{Mobile CPU (Kryo 385 and 585) and GPU (Mali-G72) performance test (ResNet-18, MobileNetV2, MnasNet1.0 with ImageNet dataset)}
\footnotesize
\centering
\resizebox{.88\textwidth}{!}{ 
\begin{tabular}{p{0.155\linewidth}|p{0.245\linewidth}|p{0.17\linewidth}|p{0.1\linewidth}p{0.1\linewidth}p{0.08\linewidth}p{0.08\linewidth}}
\hline 
\small Model & Method & \multirow{2}{\linewidth}{FPS \\ (Increase rate)} & FLOPS & Params & Top-1 Acc & Top-5 Acc \\ 
\hline
\multirow{5}{\linewidth}{ResNet-18 \\ (Kryo 385)} 
                            & Original (TVM) & 18.86 & 1.81B & 11.7M & 69.76\% & 89.08\% \\
                            & PQF~\cite{martinez2021permute}+TVM & 18.73 (0.99$\times$) & 166M & 8.26M & 66.74\% & 87.16\% \\
                            & FPGM~\cite{he2019filter}+TVM & 22.93 (1.22$\times$) & 1.10B & 7.30M & 68.37\% & 88.43\% \\
                            & NetAdapt~\cite{yang2018netadapt}+TVM & 35.17 (1.86$\times$) & 1.24B & 9.59M & 68.45\% & 88.37\% \\ 
                            & \systemName\ & 36.92 (1.96$\times$) & 1.17B & 10.3M & 68.30\% & 88.34\% \\ \hline 
\multirow{4}{\linewidth}{ResNet-18 (Mali-G72)} & Original & 15.65 & 1.81B & 11.7M & 69.76\% & 89.08\% \\
                            & PQF+TVM & 24.14 (1.54$\times$) & 166M & 8.26M & 66.74\% & 87.16\% \\
                            & FPGM+TVM & 26.62 (1.70$\times$) & 1.10B & 7.30M & 68.37\% & 88.43\% \\ 
                            & \systemName\ & 30.02 (1.92$\times$) & 1.55B & 10.4M & 69.83\% & 89.24\% \\ \hline 
\multirow{3}{\linewidth}{MobileNetV2 (Kryo 385)} & Original & 28.20 & 301M & 3.47M & 71.88\% & 90.29\% \\
                          & AMC~\cite{he2018amc}+TVM & 67.62 (2.40$\times$) & 211M & 2.31M & 70.85\% & 89.91\% \\
                          & \systemName\ & 76.92 (2.73$\times$) & 255M & 3.29M & 70.33\% & 89.57\% \\ \hline 
\multirow{3}{\linewidth}{MobileNetV2 (Mali-G72)} & Original & 68.68 & 301M & 3.47M & 71.88\% & 90.29\% \\
                          & AMC+TVM & 90.58 (1.32$\times$) & 211M & 2.31M & 70.85\% & 89.91\% \\
                          & \systemName\ & 101.56 (1.48$\times$) & 281M & 3.31M & 71.39\% & 90.16\% \\ \hline 
\multirow{2}{\linewidth}{MnasNet1.0 \\ (Kryo 585)} & Original & 42.92 & 314M & 4.35M & 73.46\% & 91.51\% \\
                          & \systemName\ & 56.85 (1.32$\times$) & 284M & 3.82M & 72.90\% & 91.16\% \\ \hline 
\end{tabular}
}
\label{tab:performance_test_imagenet}
\end{table*}

\subsection{Effect of Pruning on Associated Subgraphs} \label{associated_single}
\begin{figure}[t]
    \begin{center}
            \subfigure[Relative time cost]{
                \includegraphics[width=0.38\columnwidth]{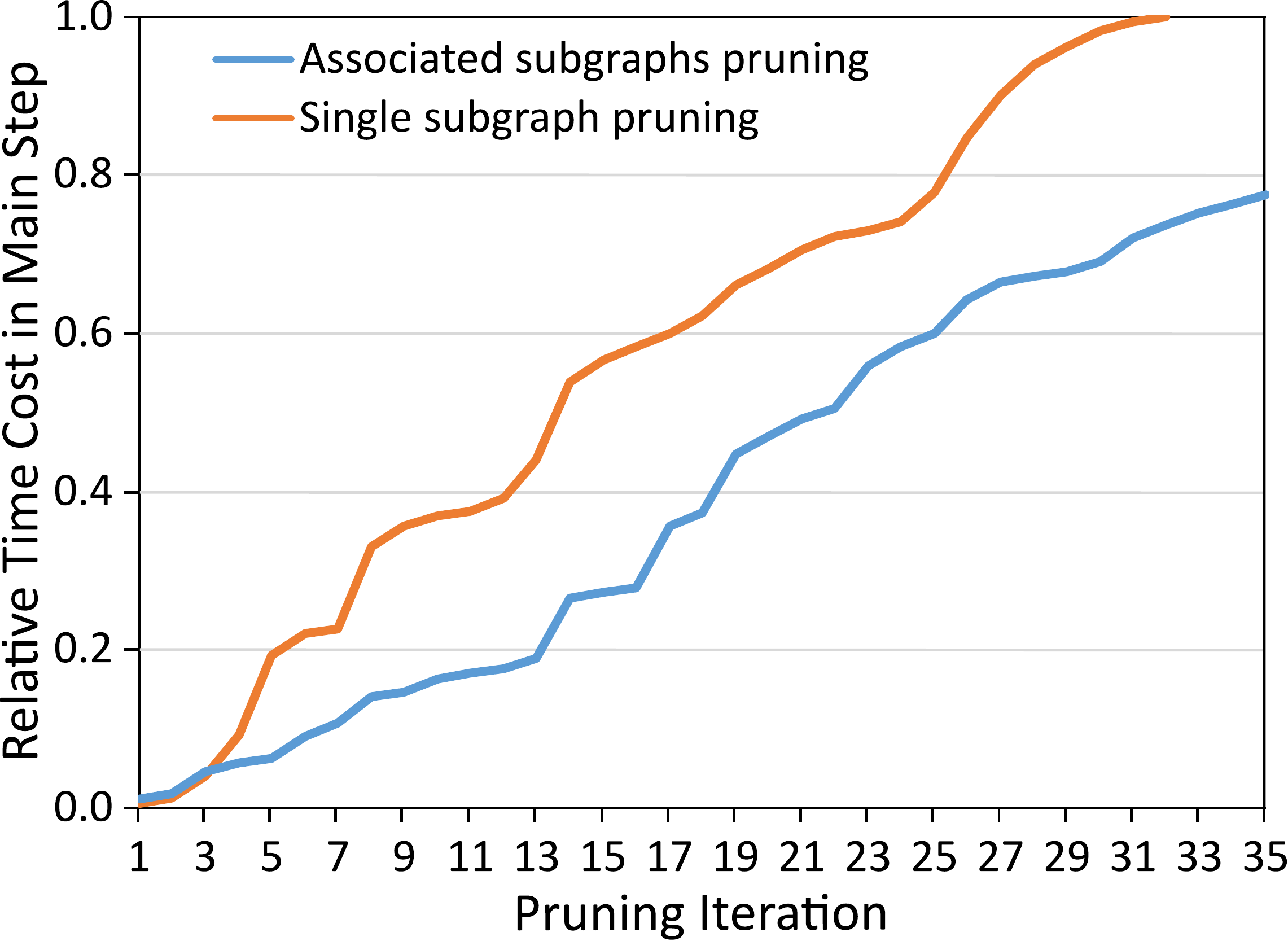}
            }
            \subfigure[FPS]{
                \includegraphics[width=0.37\columnwidth]{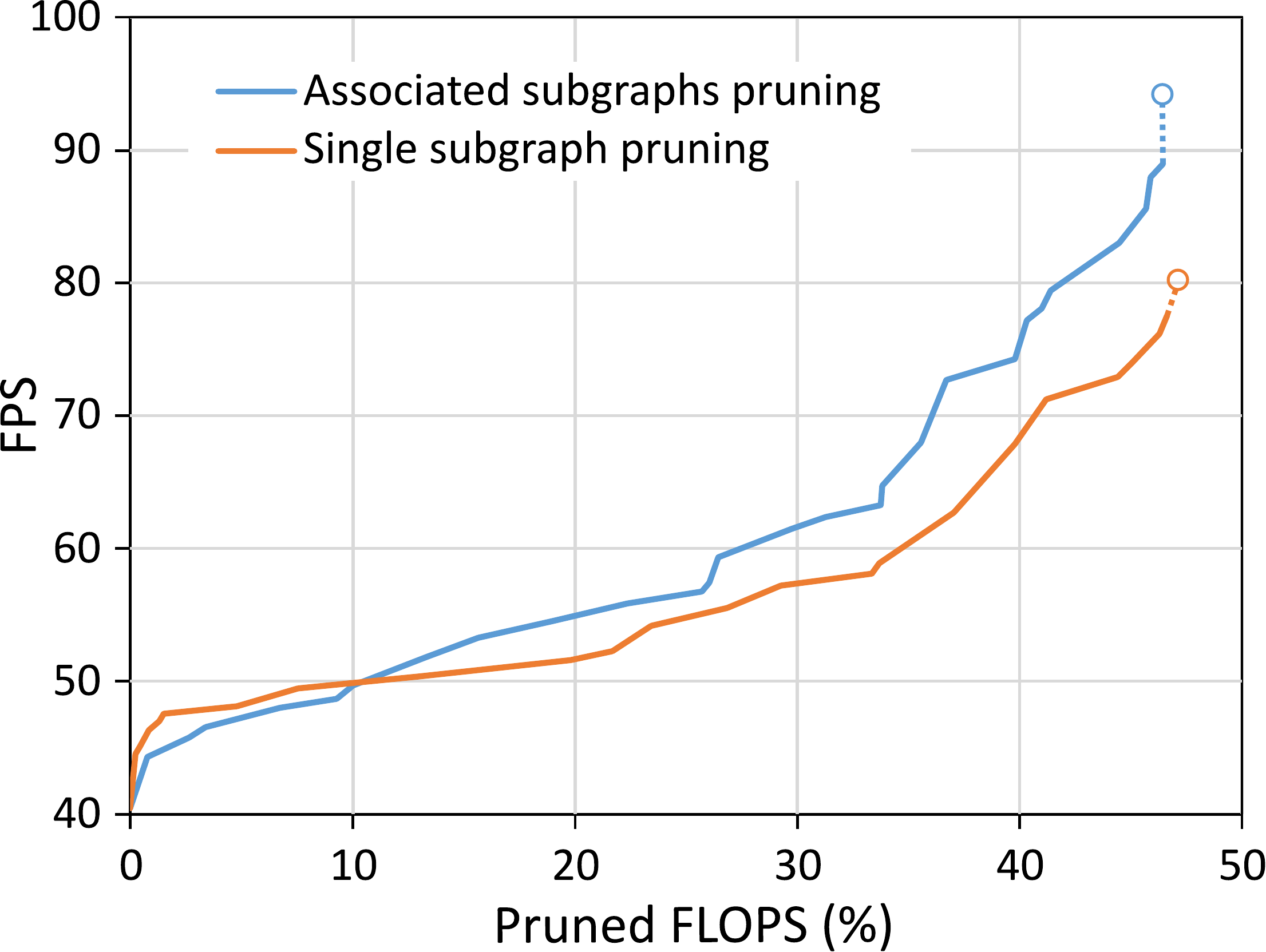}
            }
    \end{center}
    \caption{The effect of pruning all the subgraphs associated with the task (ResNet-18, Kryo 585, CIFAR-10). (a) is the relative time cost comparison with pruning a single subgraph at a time}
    \label{fig:single_associated}
\end{figure}

This experiment checks the effectiveness of pruning filters of all subgraphs related to a task as in \systemName\ than pruning filters of only one subgraph in each iteration. 
For a task associated with greater than one subgraph, we have a design choice to prune filters of a single subgraph at a time (NetAdapt~\cite{yang2018netadapt}) or prune filters of all the subgraphs associated with the task (like \systemName). 
Associated subgraphs pruning can shorten the \systemName\ process in proportional to the number of related subgraphs in the model. We observe that the associated subgraphs pruning consumes relatively less time in the Main step of \systemName\ than the single subgraph pruning, as shown in Figure~\ref{fig:single_associated} (a). In addition, the associated subgraphs pruning improves the FPS by more than 13 FPS compared to the single subgraph pruning strategy without significantly reducing the accuracy, as shown in Figure~\ref{fig:single_associated} (b) and Table~\ref{tab:performance_test_cifar10}.

\begin{table*}[t]
\caption{Mobile CPU performance test (ResNet-18 with CIFAR-10 dataset)}
\centering
\resizebox{.88\textwidth}{!}{ 
\begin{tabular}{p{0.135\linewidth}|p{0.25\linewidth}|p{0.23\linewidth}p{0.1\linewidth}p{0.1\linewidth}p{0.125\linewidth}}
\hline
\small Model & Method & FPS (Increase rate) & FLOPS & Params & Top-1 Acc \\ 
\hline
\multirow{2}{\linewidth}{ResNet-18 \\ (Kryo 280)} 
                           & Original (TVM) & 33.82 & 555M & 11.2M & 94.37\% \\
                           & \systemName\ & 109.45 (3.24$\times$) & 161M & 2.62M & 93.74\% \\ 
\hline                           
\multirow{4}{\linewidth}{ResNet-18 \\ (Kryo 585)} 
                           & Original & 40.50 & 555M & 11.2M & 94.37\% \\
                           & \systemName\ & 93.63 (2.31$\times$) & 297M & 3.54M & 94.14\% \\ 
                           & \systemName\ (w/o tuning) & 57.77 (1.43$\times$) & 390M & 5.08M & 94.51\% \\ 
                           & \systemName\ (single subgraph pruning) & 79.62 (1.97$\times$) & 294M & 4.55M & 94.27\% \\ 
\hline
\end{tabular}
}
\label{tab:performance_test_cifar10}
\end{table*}

\section{Conclusion}
Existing methods generate a compressed model by focusing on a model itself for fast DNN model execution on resource-constrained target devices. However, we have confirmed that knowing which model is best from pruning is impossible without considering compiler optimization. Therefore, we propose \systemName, a compiler-informed model pruning for efficient target-aware DNN execution. \systemName\ ensures the actual performance with pruning on the target device by using the relationship between the task and subgraphs during the compiler optimization. \systemName\ generates a target-aware DNN execution model by pruning a model based on on-device performance and compiler optimization. We verified that the pruned model generated by \systemName\ improves a DNN model's execution speed significantly while meeting the accuracy requirement.

\noindent
\textbf{Acknowledgments:} This work was supported by the Institute of Information \& communications Technology Planning \& Evaluation (IITP) grant funded by the Korea government (MSIT) (No. 2018-0-00769, Neuromorphic Computing Software Platform for Artificial Intelligence Systems).

\clearpage
%
%
\bibliographystyle{splncs04}
\bibliography{egbib}
\end{document}